\documentclass[preprint,12pt,authoryear]{elsarticle}

\usepackage{amssymb}
\usepackage{amsmath}
\usepackage{color}
\begin{document}

\begin{frontmatter}



\title{Analysis on Gradient Propagation in Batch \\
Normalized Residual Networks}


\author{Abhishek Panigrahi, Yueru Chen and C.-C. Jay Kuo \\
\texttt{\{abhishekpanigrahi034@gmail.com, yueruche@usc.edu, cckuo@sipi.usc.edu}\} 
}

\address{
Ming-Hsieh Department of Electrical Engineering\\
University of Southern California\\
Los Angleles, CA 90089, USA}

\begin{abstract}
We conduct mathematical analysis on the effect of batch normalization
(BN) on gradient backpropogation in residual network training, which is
believed to play a critical role in addressing the gradient
vanishing/explosion problem, in this work. By analyzing the mean and
variance behavior of the input and the gradient in the forward and
backward passes through the BN and residual branches, respectively, we
show that they work together to confine the gradient variance to a
certain range across residual blocks in backpropagation.  As a result,
the gradient vanishing/explosion problem is avoided. We also show the relative importance of batch normalization w.r.t. the residual branches in residual networks. %Furthermore, we
%use the same analysis to discuss the tradeoff between depths and widths
%of a residual network and demonstrate that shallower yet wider resnets
%have stronger learning performance that deeper yet thinner resnets. 
\end{abstract}


\end{frontmatter}



\section{Introduction}\label{sec:introduction}

Convolutional neural networks (CNNs) \citep{lecun1989backpropagation,
bengio2009learning, krizhevsky2012imagenet} aim at learning a feature
hierarchy where higher level features are formed by the composition of
lower level features. The deep neural networks act as stacked networks
with each layer depending on its previous layer's output.  The
stochastic gradient descent (SGD) method
\citep{simard1998transformation} has proved to be an effective way in
training deep networks.  The training proceeds in steps with SGD, where
a mini-batch from a given dataset is fed at each training step. However,
one factor that slows down the stochastic-gradient-based learning of
neural networks is the internal covariate shift.  It is defined as the
change in the distribution of network activations due to the change in
network parameters during the training. 

To improve training efficiency, \citet{ioffe2015batch} introduced a
batch normalization (BN) procedure to reduce the internal covariate
shift. The BN changes the distribution of each input element at each
layer. Let ${\bf x} =(x_{1}, x_{2}, \cdots , x_{K})$, be a K-dimensional
input to a layer.  The BN first normalizes each dimension of ${\bf x}$ as
\begin{equation}
x^{new}_{k} = \frac{x_{k} - E(x_{k})}{\sqrt{Var(x_{k})}}, 
\end{equation}
and then provide the following new input to the layer
\begin{equation}
z_{k} = \gamma_{k} x^{new}_{k} + \beta_{k}, 
\end{equation}
where $k=1, \cdots, K$ and $\gamma_{k}$ and $\beta_{k}$ are parameters
to be determined. \citet{ioffe2015batch} offered a complete analysis on
the BN effect along the forward pass. However, there was little
discussion on the BN effect on the backpropagated gradient along the
backward pass.  This was stated as an open research problem in
\citep{ioffe2015batch}. Here, to address this problem, we conduct a
mathematical analysis on gradient propagation in batch normalized
networks. 

The number of layers is an important parameter in the neural network design.
The training of deep networks has been largely addressed by normalized
initialization \citep{simard1998transformation, glo2015understanding,
saxe2013exact, he2015delving} and intermediate normalization layers
\citep{ioffe2015batch}. These techniques enable networks consisting of
tens of layers to converge using the SGD in backpropagation.  On the
other hand, it is observed that the accuracy of conventional CNNs gets
saturated and then degrades rapidly as the network layer increases. Such
degradation is not caused by over-fitting since adding more layers to a
suitably deep model often results in higher training errors
\citep{srivastava2015highway, he2015convolutional}.  To address this
issue, \citet{he2016deep} introduced the concept of residual branches.
A residual network is a stack of residual blocks, where each residual
block fits a residual mapping rather than the direct input-output
mapping. A similar network, called the highway network, was introduced
by \citet{srivastava2015highway}. Being inspired by the LSTM model
\citep{gers1999learning}, the highway network has additional gates in
the shortcut branches of each block. 

There are two major contributions in this work. First, we propose a
mathematical model to analyze the BN effect on gradient propogation in
the training of residual networks.  It is shown that residual networks
perform better than conventional neural networks because residual
branches and BN help maintain the gradient variation within a range
throughout the training process, thus stabilizing
gradient-based-learning of the network. They act as a check on the
gradients passing through the network during backpropagation so as to
avoid gradient vanishing or explosion. Second, we show that BN is vital to the training of residual networks.

The rest of this paper is organized as follows.  Related previous work
is reviewed in Sec. \ref{sec:review}. Next, we derive a mathematical
model for gradient propagation through a layer defined as a combination
of batch normalization, convolution layer and ReLU in Sec. \ref{sec:GP}.
Then, we apply this mathematical model to a resnet block in Sec.
\ref{sec:GP-Block}.  Afterwards, we experimentally show the relative importance of batch normalization w.r.t. the residual branches in residual networks in Sec. \ref{sec:batchvsres}. Concluding remarks and future
research directions are given in Sec. \ref{sec:conclusion}. 

\section{Review of Related Work}\label{sec:review}

One major obstacle to the deep neural network training is the
vanishing/exploding gradient problem \citep{bengio1994learning}. It
hampers convergence from the beginning. Furthermore, a proper
initialization of a neural network is needed for faster convergence to a
good local minimum.  \citet{simard1998transformation} proposed to
initialize weights randomly, in such a way that the sigmoid is activated in its
linear region. They implemented this choice by stating that the standard
deviation of the output of each node should be close to one. 

\citet{glo2015understanding} proposed to adopt a properly
scaled uniform distribution for initialization.  Its derivation was
based on the assumption of linear activations used in each layer . Most recently, \citet{he2015delving}
took the ReLU/PReLU activation into consideration in deriving their
proposal.  The basic principle used by both is that a proper
initialization method should avoid reducing or magnifying the magnitude
of the input and its gradient exponentially. To achieve this objective,
they first initialized weight vectors with zero mean and a certain
variance value.  Then, they derived the variance of activations at each
layer, and equated them to yield an initial value for the variance of
weight vectors at each layer. Furthermore, they derived the variance of
gradients that are backpropagated at each layer, and equated them to
obtain an initial value for the variance of weight vectors at each
layer. They either took an average of the two initialized weight
variances or simply took one of them as the initial variance of weight vectors. Being built up on this idea, we
attempt to analyze the BN effect by comparing the variance of gradients
that are backpropagated at each layer below. 

\section{Gradient Propagation Through A Layer}\label{sec:GP} 

\subsection{BN Layer Only}

We first consider the simplest case where a layer consists of the BN
operation only.  We use ${\bf x}$ and ${\tilde {\bf x}}$ to denote a
batch of input and output values to and from a batch normalized (BN)
layer, respectively.  The standard normal variate of ${\bf x}$ is ${\bf
z}$ i.e. the vector  ${\bf
z}$ is calculated by element wise normalization of input batch ${\bf x}$. In gradient backpropagation, the batch of input gradient values to
the BN layer is $\Delta{\tilde {\bf x}}$ while the batch of output
gradient values from the BN layer is $\Delta{\bf x}$.  Mathematically, we have
\begin{equation}
\tilde {\bf x} = BN ({\bf x})
\end{equation}
By simple manipulation of the formulas given in \cite{ioffe2015batch}, we 
can get 
\begin{equation}
\Delta{x_{i}} =  \frac{\gamma}{Std(x_i)}((\Delta{\tilde{x}_{i}} - E(\Delta{\tilde{x}_i})) 
- z_{i} E(\Delta{\tilde{x}_i} z_i)),
\end{equation}
where $x_{i}$ is the $i$th feature of ${\bf x}$ and $Std()$ is the 
standard deviation of element $x_{i}$ across the batch. Then, it is straightforward to derive
\begin{equation}\label{eq:BN_only}
E(\Delta {x}_i) = 0, \quad \mbox{and} \quad Var(\Delta{x}_i) = 
\frac{\gamma^2}{Var(x_i)}(Var(\Delta \tilde{x}_i ) - (E(\Delta 
\tilde{x}_i z_i))^2).
\end{equation}

%%%%%%%%%%%%%%%%%%%%%%%%%%%%%%%%%%%%%%%%%%
\begin{figure}[!ht]
\centering 
\includegraphics[width=\textwidth, height = 0.2\linewidth]{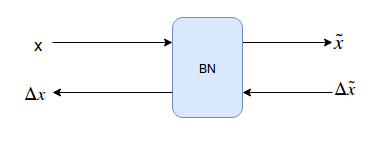}
\caption{Illustration of a BN layer.}\label{BN}
\end{figure}
%%%%%%%%%%%%%%%%%%%%%%%%%%%%%%%%%%%%%%%%%%

\subsection{Cascaded BN/ReLU/CONV Layer}

Next, we examine a more complex but common case, where a layer consists
of three operations in cascade. They are: 1) batch normalization, 2)
ReLU activation, and 3) convolution. Here, we take BN and ReLU before convolution because we want to explore the activation of a convolution layer by taking the layer input activation into consideration. It doesn't matter whether BN and ReLU are actually placed before or after convolution, because if they are actually placed after a convolution layer, we can consider our calculations taking them as placed before the next convolution layer. To simplify the gradient flow calculation, we make some assumptions which will be mentioned whenever needed. 

The input to the $L$th Layer of a deep neural network is ${\bf y}_{L-1}$
while its output is ${\bf y}_L$.  We use $BN$, $ReLU$ and $CONV$ to
denote the three operations in each sub-layer. Then, we have the
following three equations:
\begin{equation}
\tilde{\bf y}_{L-1} = BN ({\bf y}_{L-1}), \quad
\hat{\bf y}_{L-1} = ReLU (\tilde{\bf y}_{L-1}), \quad
{\bf y_L} = CONV (\hat{\bf y}_{L-1}).
\end{equation}

\begin{figure}[!ht]
\centering 
\includegraphics[width=\textwidth, height = 0.2\linewidth]{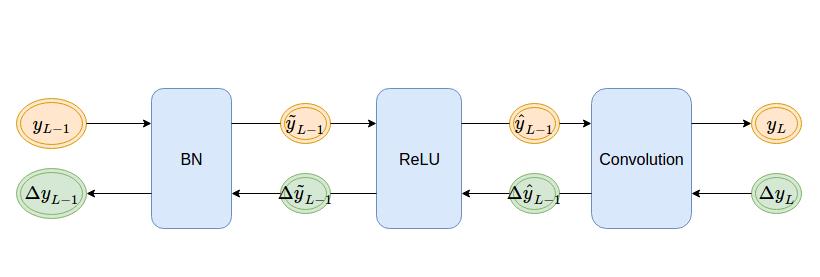}
\caption{Illustration of a layer that consists of BN, ReLU and CONV three
sub-layers.}\label{cascade-layer}
\end{figure}

The relationship between ${\bf y}_{L-1}$, $\tilde{\bf y}_{L-1}$,
$\hat{\bf y}_{L-1}$ and ${\bf y}_L$ is shown in Fig.
\ref{cascade-layer}.  As shown in the figure, $\tilde{\bf y}_{L-1}$
denotes the batch of output elements from the BN sub-layer. It also
serves as the input to the ReLU sub-layer. $\hat{\bf y}_{L-1}$ denotes
the batch of output elements from the ReLU sub-layer. It is fed into the
convolution sub-layer. Finally, ${\bf y}_{L}$ is the batch of output
elements from the CONV sub-layer. Gradient vectors have $\Delta$ as the
prefix to their corresponding vectors in the forward pass.  In this figure, ${\bf W}_{L}$ is the weight vector of the convolution layer. The dimensions of ${\bf y}_{L}$ and $\Delta{\bf y}_{L}$ are $n_{L}$ and $n_{L}^{\prime}$, respectively.  $y_{L-1,i}$ denotes the $i$th feature
of activation ${\bf y}_{L-1}$. %The expectation $E(v_{i})$ denotes expectation of the $i_{th}$ element of variable v across its batch, while E(v) denotes expectation taken over the entire variable v. 

%The variance and mean of any activation or gradient is always calculated across a batch of activations or gradients because we adopt the batch as a representative of the entire sample. 

%%%%%%%%%%%%%%%%%%%%%%%%%%%%%%%
%%%%%%%%%%%%%%%%%%%%%%%%%%%%%%%
Please note that from now on in the derived equations, $Var(y_{L})$ denotes a vector, where each element $Var(y_{L,i})$ denotes the variance of element $y_{L,i}$ across its batch. $Var(W_{L, .})$ denotes the variance of the entire weight matrix. To simplify representation, we denote $W^{2}$ as the element wise squared matrix of W. Also, $W^T$ denotes the transpose of matrix W.

\subsubsection{Variance Analysis in Forward Pass}

We will derive the mean and variance of output $y_{L,i}$ from the input
${\bf y}_{L-1}$.  First, we examine the effect of the BN sub-layer. The
output of a batch normalization layer is $\gamma_i z_i + \beta_i$, where
$z_i$ is the standard normal variate of $y_{L-1,i}$, calculated across a 
batch of activations. Clearly, we have
\begin{equation}
E(\tilde{y}_{L-1,i}) = \beta_i, \quad \mbox{and} \quad
Var(\tilde{y}_{L-1,i}) = \gamma_i^2.
\end{equation}
Next, we consider the effect of the ReLU sub-layer.  Let
$a=\frac{\beta_i}{\gamma_i}$. \iffalse
%%%%%%%%%%%%%%%%%%%%%%%%%%%%%%%%%%
We assume that $a$ is small enough so that
the standard normal variate $z_i$ follows a nearly uniform distribution
in interval $(0, a)$. 
%%%%%%%%%%%%%%%%%%%%%%%%%%%%%%%%%%
\fi 
In Appendix A, we show a step-by-step procedure
to derive the mean and variance of the output of the ReLU sub-layer when
it is applied to the output of a BN layer. Here, we summarize the main
results below:
%\color{magenta}
\begin{eqnarray}
E(\hat{y}_{L-1,i}) & = & \gamma_i  \left(\frac{1}{\sqrt{2\pi}} + \frac{a}{2} + \frac{1}{\sqrt{2 \pi}} (1 - \exp(\frac{-a^2}{2})) \right) \\
E(\hat{y}_{L-1,i}^{2}) & = & 
E(y^2) = 0.5 + \sqrt{\frac{2}{\pi}} a + 0.5 a^2+\exp(\frac{-a^2}{2}) + p(a)
\end{eqnarray}

where p(a) =  $\int_{0}^{a} \frac{1}{\sqrt{2 \pi}} \exp(\frac{-z^2}{2})dz$.

\color{black}
%%%%%%%%%%%%%%%%%%%%%%%%%%%%%%%%%%%%%%%%%%%%%
\iffalse
Finally, we consider the influence of the CONV sub-layer. To simplify
the analysis, we assume that all elements in ${\bf W}_{L}^f$ are
mutually independent and with the same distribution of mean 0 and all elements in
${\bf y}_{L-1}$ are also mutually independent and with the same
distribution across a batch of activations. Furthermore, ${\bf y}_{L-1}$
and ${\bf W}_{L}^f$ are independent of each other. Then, we get
\begin{equation}\label{eq:variance}
Var(y_{L,i}) = n_{L} Var(W_{L,i}^f) E((\hat{y}_{L-1,i})^2).
\end{equation}
\fi
%%%%%%%%%%%%%%%%%%%%%%%%%%%%%%%%%%%%%%%%%%%%
Finally, we consider the influence of the CONV sub-layer. $W_{L}$ is the matrix of the CONV sub-layer of dimension ($n_{L}^{\prime}, n_{L}$) i.e. $(y_{L})_{n_{L}^{\prime} * 1}$ = $(W_{L})_{n_{L}^{\prime} * n_{L}}$ $(\hat{y}_{L-1})_{ n_{L} * 1}$. Here, we assume that all elements in ${\bf \hat{y}}_{L-1}$ are mutually independent. Then, it's trivial to see that since we are calculating the variance across a batch of activations,
\begin{equation}\label{eq:variance}
Var(y_{L}) = W_{L}^{2} Var(\hat{y}_{L-1})
\end{equation}

\iffalse
Note that assuming the weight elements come from a distribution with mean 0 is a fair assumption because we initialize the weight elements from a distribution with mean 0 and in the next subsection, we see that the mean of gradient that reaches the convolution layer during backpropagation has mean 0 across a batch.
If y = W x, where y and x are m dimensional and k dimensional vectors respectively, and W is (m, k) dimensional matrix, then $\Delta{w_{i,j}}$ = $\sum_{batch}$ $x_{i}$ $\Delta{y_{j}}$. Since, the distribution of an element of gradient vector y is of zero mean across a batch, we can fairly assume the final distribution of weight gradient E($\Delta{w_{i,j}}$) as 0.
\fi

\subsubsection{Variance Analysis in Backward Pass}

We consider backward propagation from the $L$th layer to the $(L-1)$th
layer and focus on gradient propagation. Since, the gradient has just
passed through the BN sub-layer of Lth layer, using eq. (\ref{eq:BN_only}) we get E($\Delta{\bf y}_{L}$) = 0. Note that here 0 denotes the vector 0 in dimension of length of $y_{L}$. 

%%%%%%%%%%%%%%%%%%%%%%%%%%%%%%%%%%%%%%%%%%%%%
\iffalse
Under the following three assumptions: 1) elements in ${\bf W}_{L}^{b}$
are mutually independent and with the same distribution of mean 0, 2) elements in
$\Delta{\bf y}_{L}$ are mutually independent and with the same
distribution across a batch, and 3) $\Delta{\bf y}_{L}$ and ${\bf
W}_{L}^{b}$ are independent of each other.  Then, we get
\begin{equation}
Var(\Delta \hat y_{L-1,i}) = n_{L}^{\prime} Var(\Delta y_{L,i}) Var(W_{L,i}^{b}), 
\mbox{ and }\\ E(\Delta \hat y_{L-1,i}) = E(\Delta{\bf y}_{L,i}) = 0.
\end{equation}
\fi
%%%%%%%%%%%%%%%%%%%%%%%%%%%%%%%%%%%%%%%%%%%%%

First, gradients go through the CONV sub-layer. $(\Delta{\hat{y}}_{L-1})_{n_{L}*1}$ = $(W_{L}^{T})_{n_{L}*n_{L}^{\prime}}$ $(\Delta{y}_{L})_{n_{L}^{\prime}*1}$. Here, we assume that all elements in ${\bf \Delta{y}}_{L}$ are mutually independent. Then, it's trivial to see that since we are calculating the expectation and variance across a batch of activations,

\begin{equation}
E(\Delta{\hat{y}}_{L-1}) =  W_{L}^{T} E(\Delta{\bf y}_{L}) = 0 \mbox{ and }
Var(\Delta{\hat{y}}_{L-1}) = (W_{L}^{T})^2 Var(\Delta{y}_{L}) 
\end{equation} 

Next, gradients go through the ReLU sub-layer. It is assumed that the
function applied to the gradient vector on passing through ReLU and the
elements of gradient are independent of each other. Since the input in
the forward pass was a shifted normal variate ($a=\frac{\beta_i}{\gamma_i}$), 
we get

%\color{magenta}
\begin{equation}\label{eq:Relu_grad}
\begin{aligned}
&E(\Delta \tilde{y}_{L-1,i}) =  (0.5 + p(a)) 
E(\Delta {\hat{y}_{L-1,i}}) = 0.0, \mbox{ and }\\
&Var(\Delta \tilde{y}_{L-1,i}) = (0.5 + p(a)) 
Var(\Delta {\hat{y}_{L-1,i}}).
\end{aligned}
\end{equation}

where p(a) =  $\int_{0}^{a} \frac{1}{\sqrt{2 \pi}} \exp(\frac{-z^2}{2})dz$.

\color{black}
In the final step, gradients go through the BN sub-layer. If the
standard normal variate, ${\bf z}$, to the BN sub-layer and the incoming
gradients $\Delta {\bf y}$ are independent, we have $E(z_i \Delta
y_{L-1,i})= E(z_i) E(\Delta y_{L-1,i}) = 0$. The last equality holds since the mean of the standard normal variate is zero.

The final result is
%\color{magenta}
\begin{equation}\label{eq:gradient}
Var(\Delta y_{L-1, i}) = \frac{\sum_{j=1}^{n_{L}^{\prime}} W_{L, ji}^{2} Var(\Delta y_{L, j})}{\sum_{j=1}^{n_{L-1}} W_{L-1, ij}^{2}}\frac{0.5+p(a)}{0.5 + \sqrt{\frac{2}{\pi}} a + 0.5 a^2+\exp(\frac{-a^2}{2}) + p(a)} 
\end{equation}.

where p(a) = $\int_{0}^{a} \frac{1}{\sqrt{2 \pi}} \exp(\frac{-z^2}{2})dz$

\color{black}
Let's see what the above equation means. The numerator shows a weighted sum of the gradient elements of Lth layer, the weights being the ith column of weight matrix $W_{L}$. While the denominator shows a simple summation of the ith row of weight matrix $W_{L-1}$. Now, we take some assumptions to simplify the above equation, derive some meaning of the above equation and find the expectation of gradient vector $y_{L}$. To simplify the analysis, we assume that all elements in ${\bf W}_{L}$ are of the same distribution of mean 0. All elements in Var($\Delta y_{L-1}$) are from the same distribution. Furthermore, Var($\Delta y_{L-1}$)
and ${\bf W}_{L}$ are independent of each other. Also, I assume that the weight variables are bounded i.e. they don't increase indefinitely. This is a fair assumption and this is required to show that 
$$
E(\frac{1}{X}) E(X) \leq \frac{(c+d)^2}{4cd}
$$
where X is a variable and lies in the range (c,d), $0 < c < d$.  Also, for the same variable X, since $\frac{1}{X}$ is convex function, 
$$
E(\frac{1}{X}) E(X) \geq 1
$$

Using the above properties and assumptions, we get, 
$$
\begin{aligned}
\frac{n_{L}^{\prime}}{n_{L-1}}  
\frac{Var(W_{L})}{Var(W_{L-1})} E(Var(\Delta y_{L, i})) 
&\leq  E(\frac{\sum_{j=1}^{n_{L}^{\prime}} W_{L, ji}^{2} Var(\Delta y_{L, j})}{\sum_{j=1}^{n_{L-1}} W_{L-1, ij}^{2}}) \\ &\leq  K \frac{n_{L}^{\prime}}{n_{L-1}}  
\frac{Var(W_{L})}{Var(W_{L-1})} E(Var(\Delta y_{L, i}))
\end{aligned}
$$
where we assume that K is a constant such that when X = $\sum_{j=1}^{n_{L-1}} W_{L-1, ij}^{2}$ i.e. sum of a row of matrix $W_{L-1}^2$, the upper bound on $E(\frac{1}{X}) \leq \frac{K}{E(X)}$ holds. Thus, 

\begin{equation}\label{eq:upperbound}
\begin{aligned}
&E(Var(\Delta y_{L, i})) \frac{n_{L}^{\prime}}{n_{L-1}}\frac{Var(W_{L, .})}{Var(W_{L-1, .})} \frac{0.5+p(a)}{0.5 + \sqrt{\frac{2}{\pi}} a + 0.5 a^2+\exp(\frac{-a^2}{2}) + p(a)} 
\leq  E(Var(\Delta y_{L-1, i})) \\ &\leq  K  E(Var(\Delta y_{L, i})) \frac{n_{L}^{\prime}}{n_{L-1}}\frac{Var(W_{L, .})}{Var(W_{L-1, .})}\frac{0.5+p(a)}{0.5 + \sqrt{\frac{2}{\pi}} a + 0.5 a^2+\exp(\frac{-a^2}{2}) + p(a)}
\end{aligned}
\end{equation}

where p(a) = $\int_{0}^{a} \frac{1}{\sqrt{2 \pi}} \exp(\frac{-z^2}{2})$.

%%%%%%%%%%%%%%%%%%%%%%%%%%%%%%%%%%%%%%%%%%%%%
\iffalse
Var(\Delta y_{L-1, i})
Var(\Delta y_{L-1,i}) = Var(\Delta y_{L,i}) \frac{n_{L}^{\prime}}{n_{L-1}}  
\frac{Var(W_{L,i}^{b})}{Var(W_{L-1,i}^{f})} 
\frac{0.5+\sqrt{\frac{1}{2\pi}}a}{0.5 + \sqrt{\frac{2}{\pi}} a + 0.5 a^2
+\frac{a^3}{3\sqrt{2\pi}}}.
\fi
%%%%%%%%%%%%%%%%%%%%%%%%%%%%%%%%%%%%%%%%%%%%%

Note that the last product term in the upper and lower bound is a constant term. The other two
fractions are properties of the network, that compare two adjacent
Layers.  Also, assuming that the weights come from a distribution of mean 0 is a valid assumption because a) the weights are initialized with 0 mean and b) the gradients that come to the convolution layer have mean 0 by eq (\ref{eq:BN_only}) across batch. The skipped steps are given in Appendix B.

\subsection{Discussion}

\begin{figure*}[!htb]
\includegraphics[width=0.32\linewidth]{var_25000.png}\hfill
\includegraphics[width=0.32\linewidth]{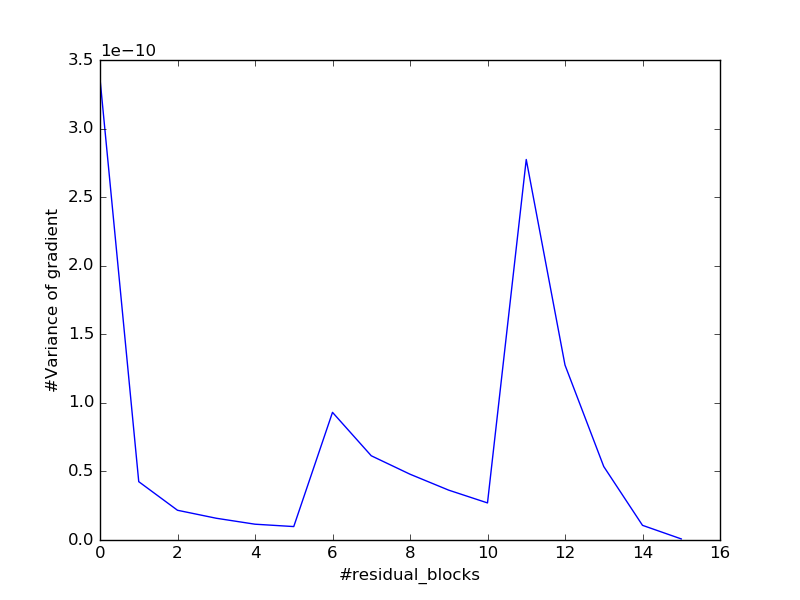}\hfill
\includegraphics[width=0.32\linewidth]{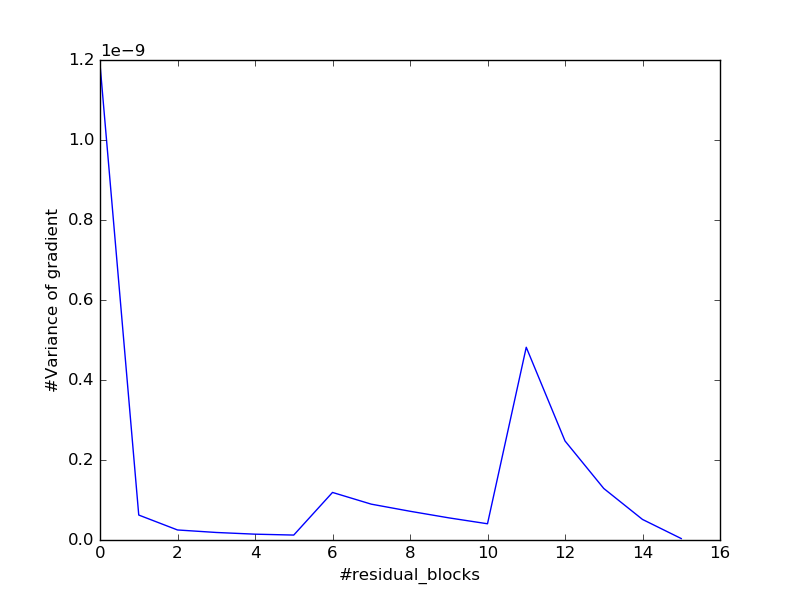}
\caption{The variance of gradient as a function of residual block measured at step 25000, when batch size is 128 (left), 256 (middle) and 512 (right) respectively.  }\label{fig:plot_165}
\end{figure*}

Initially, we set $\beta_i = 0$ and $\gamma_i = 1$ so that $a=0$.  Then,
the constant term in the RHS of Eq. (\ref{eq:gradient}) is equal to
one. Hence, if the weight initialization stays equal across all the
layers, propagated gradients are maintained throughout the network. In
other words, the BN simplifies the weight initialization job. For
intermediate steps, we can estimate the gradient variance under
simplifying assumptions that offer a simple minded view of gradient
propagation.  Note that,  when $a = \frac{\beta}{\gamma}$ is small (the experimental values  of a in fig \ref{fig:plot_165} show that it reaches at most 1.0), the
constant term is a reasonable constant (at a=1.0, the constant is around 0.33 and as the value of a decreases to 0.0, the constant approaches 1.0). \color{black} The major implication
is that, the BN helps maintain gradients across the network,
throughout the training, thus stabilizing optimization. 

\section{Gradient Propagation Through A Resnet Block}\label{sec:GP-Block} 

\subsection{Resnet Block}

The resnet blocks in the forward pass and in the gradient backpropagation pass are shown in Figs. \ref{fig:Forward} and \ref{fig:Backward}, respectively.  A residual network has multiple
scales, each scale has a fixed number of residual blocks, and the convolutional layer in residual blocks at the same scale have the same number of filters. In the analysis, we adopt the model where the filter number increases $k$ times from one scale to the next one.  Although no bottleneck blocks are explicitly considered here, our analysis holds for
bottleneck blocks as well. As shown in Fig. \ref{fig:Forward}, the
input passes through a sequence of BN, ReLU and CONV sub-layers along
the shortcut branch in the first residual block of a scale, which shapes the
input to the required number of channels in the current scale. For all
other residual blocks in the same scale, the input just passes through the shortcut branch.
For all residual blocks, the input goes through the convolution branch which consists of two sequences of BN, ReLU and CONV
sub-layers. We use a {\em layer} to denote a sequence of BN, ReLU and
CONV sub-layers as used in the last section and $F$ to denote the compound function of one layer. Note that in the first residual block, we use an explicit BN sub layer in the shortcut branch. However, in the current resnet models, the BN sublayer and ReLU sublayer are used before the residual block. The calculations won't change, if we use individual BN+ReLU in each branch but the representation is simple. 

%%%%%%%%%%%%%%%%%%%%%%%%%%%%%%%%%%%%%%%%%%%%%
\iffalse
\begin{figure}[!htb]
    \centering
    \begin{minipage}{.5\textwidth}
        \centering
        \includegraphics[width=0.8\linewidth, height=0.4\textheight]{Forward.jpg}
        \caption{A residual block in the forward pass.}
        \label{fig:Forward}
    \end{minipage}%
    \begin{minipage}{0.5\textwidth}
        \centering
        \includegraphics[width=0.8\linewidth, height=0.4\textheight]{Backward.jpg}
        \caption{A residual block in the gradient backpropagation pass.}
        \label{fig:Backward}
    \end{minipage}
\end{figure}
\fi

\begin{figure}[!ht]
\centering 
\includegraphics[width=0.8\textwidth, height = 0.3\linewidth]{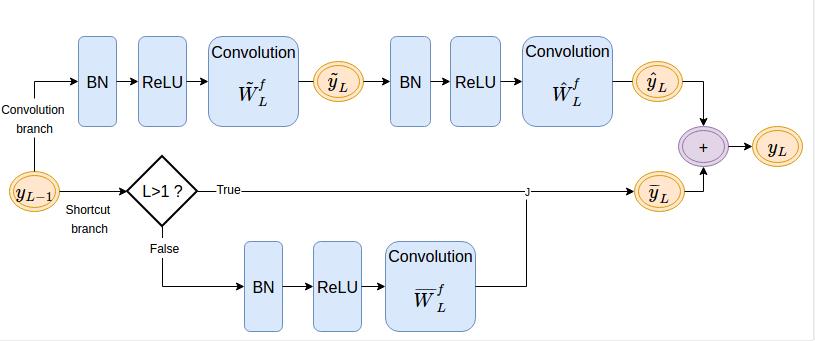}
\caption{A residual block in the forward pass.}
\label{fig:Forward}
\end{figure}

\begin{figure}[!ht]
\centering 
\includegraphics[width=0.8\textwidth, height = 0.3\linewidth]{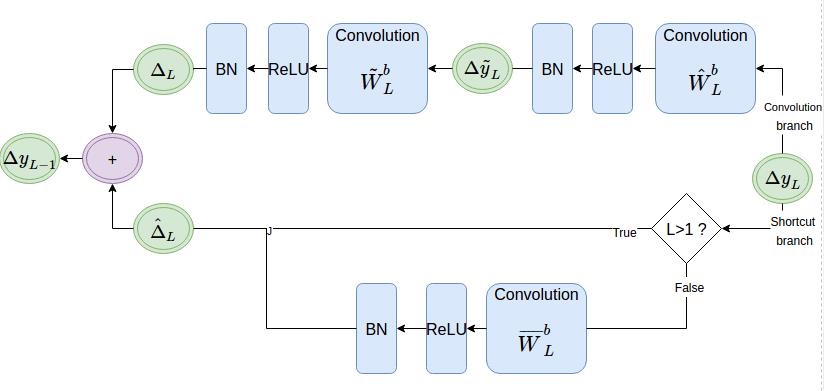}
\caption{A residual block in the gradient backpropagation pass.}
\label{fig:Backward}
\end{figure}

%%%%%%%%%%%%%%%%%%%%%%%%%%%%%%%%%%%%%%%%%%%%%

To simplify the computation of the mean and variance of $y_{L, i}$ and
$\Delta{y_{L, i}}$, we assume that $a=\frac{\beta_{i}}{\gamma_{i}}$. is a constant. We can always take  a as the average across the layers. We
define the following two associated constants.
%\color{magenta}
\begin{eqnarray}
c_1 & = & 0.5 + p(a) \label{eq:c_1} \\
c_2 & = & 0.5 + \sqrt{\frac{2}{\pi}} a + 0.5 a^2 + p(a) + exp(\frac{-a^2}{2})
\label{eq:c_2} 
\end{eqnarray}

where p(a) = where p(a) = $\int_{0}^{a} \frac{1}{\sqrt{2 \pi}} \exp(\frac{-z^2}{2})dz$.
\color{black}

which will be needed later. We show the results for the upper bound. The lower bound is just a factor smaller than the upper bound.

\subsection{Variance Analysis}

As shown in Fig. \ref{fig:Forward}, block $L$ is the $L$th residual
block in a scale with its input $y_{L-1}$ and output $y_{L}$.  The
outputs of the first and the second BN-ReLU-CONV layers in the convolution
branch are $\tilde{y}_{L} = F( y_{L-1})$ and $\hat{y}_{L} =
F(F(y_{L-1}))$, respectively.  The weight vectors of the CONV sub-layer
of the first and the second layers in the convolution branch of block $L$
are $\tilde{W}_{L}$ and $ \hat{W}_{L}$, respectively. The weight vector
in the shortcut branch of the first block is $\overline{W}_{1}$.  The output of
the shortcut branch is $\overline{y}_{L}$.  For $L=1$, we have
$\overline{y}_{1} = F(y_{0})$, where $y_{0}$ is the output of last
residual block of the previous scale.  For L$>$1, we have
$\overline{y}_{L} = y_{L-1}$.  For the final output, we have
\begin{equation}
y_{L} = \overline{y}_{L} + \hat{y}_{L}.
\end{equation}
For L$>$1, block $L$ receives an input of size $n_{s}$ in the forward
pass and an input gradient of size $n_{s}^{\prime}$ in the
backpropagation pass. Since block $1$ receives its input $y_{0}$ from
the previous scale, it receives an input of size $\frac{n_{s}}{k}$ in
the forward pass.  

By assuming $\overline{y}_{L}$ and $\hat{y}_{L}$ are independent, we 
have
\begin{equation}
Var(y_{L,i}) = Var(\overline{y}_{L,i}) + Var(\hat{y}_{L,i}). 
\end{equation}
We will show how to compute the variance of $y_{L ,i}$ step by step in
Appendix C for $L=1,\cdots, N$. When $L=N$, we obtain 
\begin{equation}
Var(y_{N}) = c_2 (\sum_{J=2}^{N} \hat{W}_{J}^{2} I_{n_J} +  \frac{1}{k} ( \overline{W}_{1}^{2} I_{n_1} + \hat{W}_{1}^{2} I_{n_1} ))%n_{s} (\sum_{J=2}^{N} Var(\hat{W}_{J,i}^{f}) + 
%\frac{1}{k} ( Var(\overline{W}_{1, i}^{f}) + Var(\hat{W}_{1, i}^{f}))),
\end{equation}
where $c_2$ is defined in Eq. (\ref{eq:c_2}) and $I_x$ denotes a one vector of dimension x.

We use $\Delta$ as prefix in front of vector representations at the corresponding positions in forward pass to denote the gradient in Fig. \ref{fig:Backward} in the
backward gradient propagation. Also, as shown in Fig. \ref{fig:Backward}, we represent the gradient vector at the tip of the convolution branch and shortcut branch by $\Delta_{L}$ and $\hat{\Delta}_{L}$ respectively. As shown in the figure, we have
\begin{equation}
\Delta y_{L-1} = \hat{\Delta}_{L} + \Delta_{L}
\end{equation}
A step-by-step procedure in computing the variance of $\Delta{y_{L-1, i}}$
is given in Appendix D. Here, we show the final result below:
\begin{equation}\label{eq:vardeltay}
\begin{aligned}
E(Var(\Delta{y_{L-1,i}})) \leq K_{L} & \left( 1 + (\frac{c_1}{c_2})^2 
\frac{Var(\tilde{W}_{L, .})}{\sum_{J=2}^{L-1} 
Var(\hat{W}_{J, .}) +  \frac{1}{k} ( Var(\overline{W}_{1, .}) 
+ Var(\hat{W}_{1, .}) )} \right) \\& E(Var(\Delta{y_{L, i}})).
\end{aligned}
\end{equation}
where $K_{L}$ denotes the necessary bound of the convolution layers in residual block L, as used in eq(\ref{eq:upperbound}).

\subsection{Discussion}

We can draw two major conclusions from the analysis conducted above.
First, it is proper to relate the above variance analysis to the
gradient vanishing and explosion problem. The gradients go through a BN
sub-layer in one residual block before moving to the next residual
block.  As proved in Sec. \ref{sec:GP}, the gradient mean is zero when
it goes through a BN sub-layer and it still stays at zero after passing
through a residual block. Thus, if it is normally distributed, the
probability of the gradient values between $\pm$ 3 standard deviations
is 99.7\%.  A smaller variance would mean lower gradient values. In
contrast, a higher variance implies a higher likelihood of
discriminatory gradients. Thus, we take the gradient variance across a
batch as a measure for stability of gradient backpropagation. 

Second, recall that the number of filters in each convolution layer of a
scale increases by $k$ times with respect to its previous scale.
Typically, $k=1$ or $2$. Without loss of generality, we can assume the
following: the variance of weights is about equal across layers,
${c_1}/{c_2}$ is a reasonably small constant, \color{black} and $k=2$. Then, Eq. (\ref{eq:vardeltay}) can
be simplified to
\begin{equation}\label{eq:vardeltay2}
Upperbound(E(Var(\Delta{y_{L-1,i}}))) \propto  K_{L} \frac{L}{L - 1} E(Var(\Delta{y_{L,i}})). 
\end{equation}
We see from above that the change in the gradient variance from one
residual block to its next is small. This is especially true when the
$L$ value is high. Thus, the gradient variance increases as we move across 
a scale. This observation can be used to explain the iterative estimation given in \cite{DBLP:journals/corr/GreffSS16}. 
The gradient variance is high in the lower blocks in a scale, while it is low in the 
upper blocks. That shows a vigorous change in weights in the lower blocks and the weight 
change gets finer as we move forward. Hence, the weights in the upper blocks are smoothly refined so that the features learned in the lower blocks gets finer as we move forward towards the upper blocks of a scale.

\subsection{Experimental Verification}

\begin{figure*}[h!]
\centering
\includegraphics[width=0.3\textwidth]{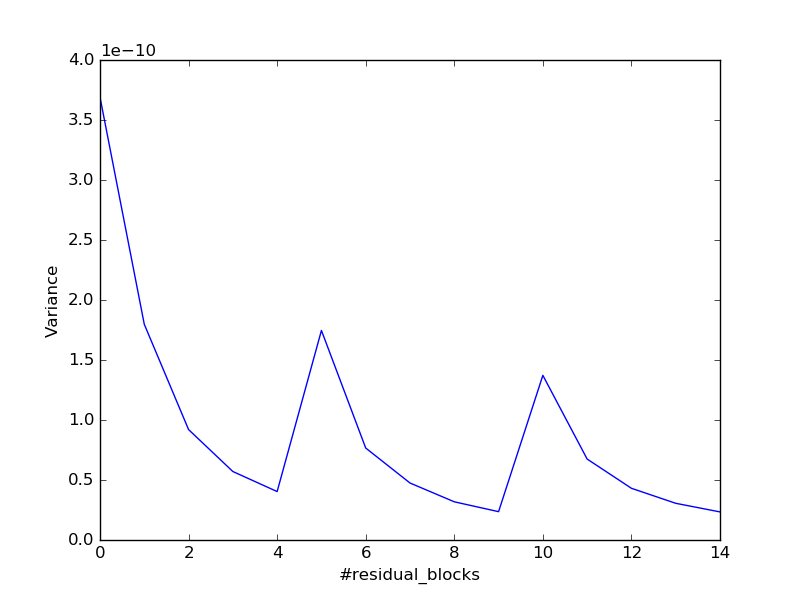}
\includegraphics[width=0.3\textwidth]{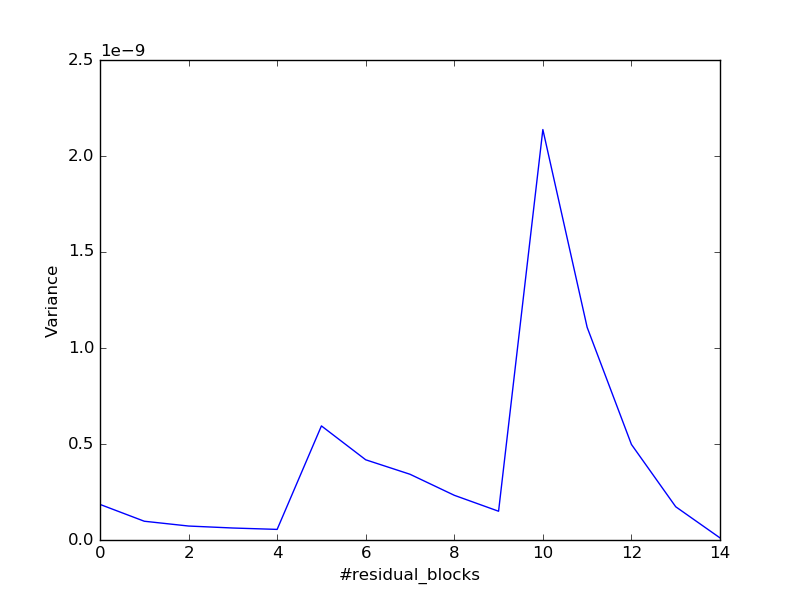}
\includegraphics[width=0.3\textwidth]{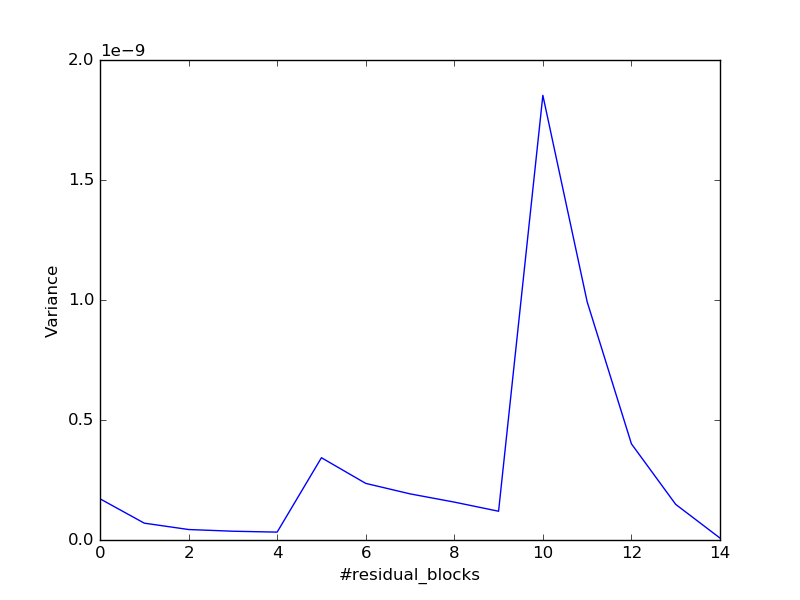}
\caption{The mean of the gradient variance as a function of the residual
block position at Epoch 1 (left), Epoch 25000 (middle) and Epoch 50000
(right).}\label{fig:plot1}
\end{figure*}

\begin{figure*}[h!]
\centering
\includegraphics[width=.3\textwidth]{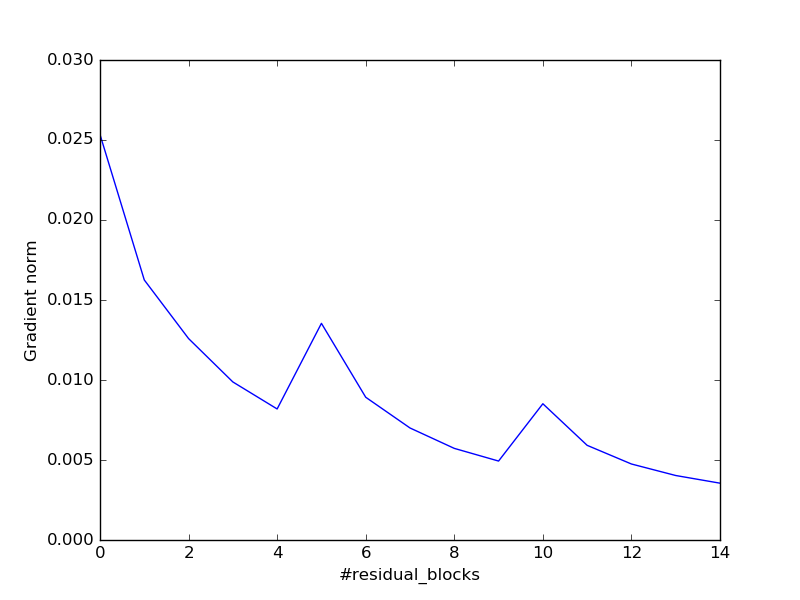}
\includegraphics[width=.3\textwidth]{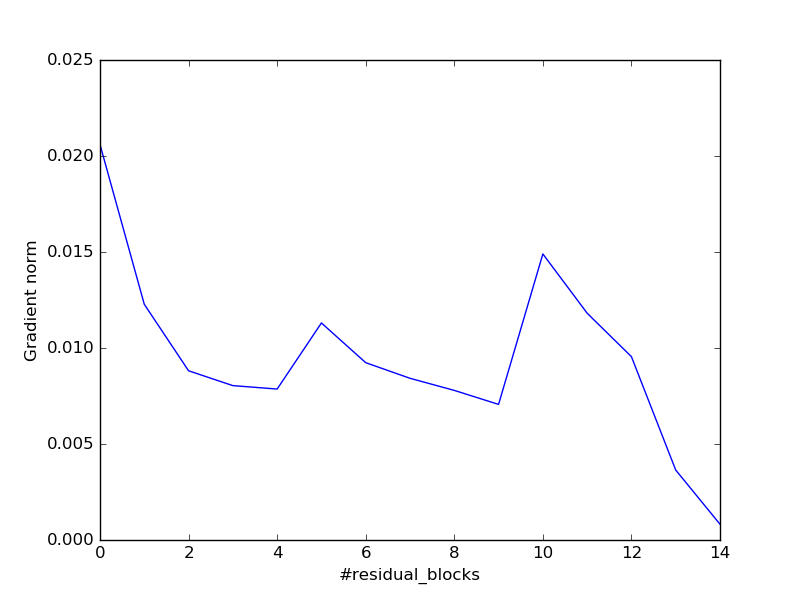}
\includegraphics[width=.3\textwidth]{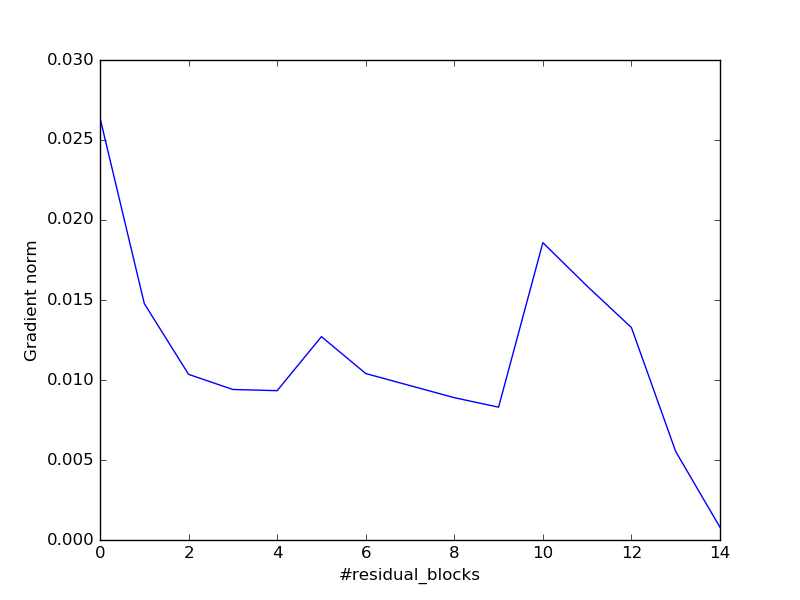}
\caption{The $l_2$ norm of the gradient as a function of the
residual block position at Epoch 1 (left), Epoch 25000 (middle) and
Epoch 50000 (right).}\label{fig:plot2}
\end{figure*}

We trained a Resnet-15 model that consists of 15 residual blocks and 3
scales on the CIFAR-10 dataset, and checked the gradient variance across
the network throughout the training. We plot the mean of the gradient variance and the $l_2$-norm of the gradient at various residual block locations in Figs.
\ref{fig:plot1} and \ref{fig:plot2}, respectively, where the gradient
variance is calculated for each feature across one batch. Since
gradients backpropagate from the output layer to the input layer, we
should read each plot from right to left to see the backpropagation
effect. The behavior is consistent with our analysis. There is a gradual
increase of the slope across a scale. The increase in the gradient
variance between two residual blocks across a scale is inversely
proportional to the distance between the residual blocks and the first
residual block in the scale. Also, there is a dip in the gradient
variance value when we move from one scale to its previous scale. Since
the BN sub-layer is used in the shortcut branch of the first block of a
scale, it ensures the decrease of the gradient variance as it goes from
one scale to another. 

%%%%%%%%%%%%%%%%%%%%%%%%%%%%%%%%%%%%%%%%%%%%%
%%%%%%%%%%%%%%%%%%%%%%%%%%%%%%%%%%%%%%%%%%%%%

%%%%%%%%%%%%%%%%%%%%%%%%%%%%%%%%%%%%%%%%%%%%%
%%%%%%%%%%%%%%%%%%%%%%%%%%%%%%%%%%%%%%%%%%%%%

\begin{figure*}[!htb]
\includegraphics[width=0.32\linewidth]{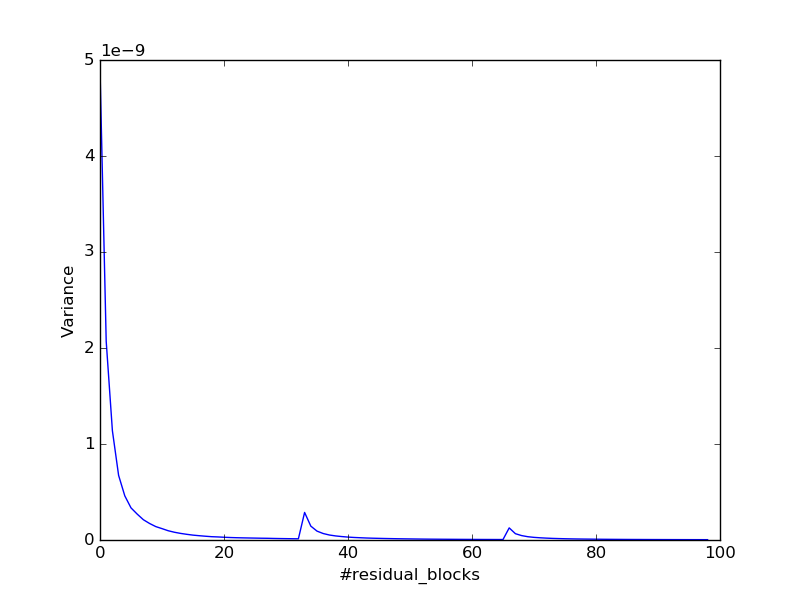}
\includegraphics[width=0.32\linewidth]{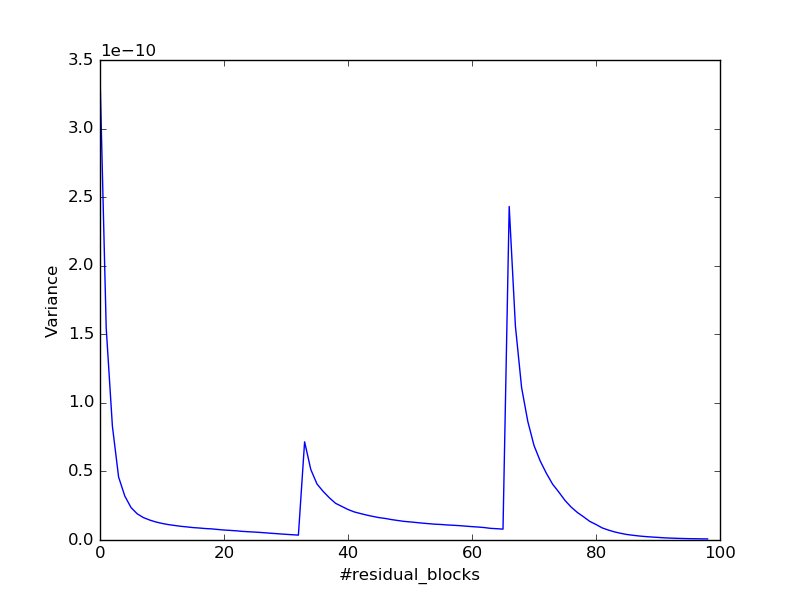}
\includegraphics[width=0.32\linewidth]{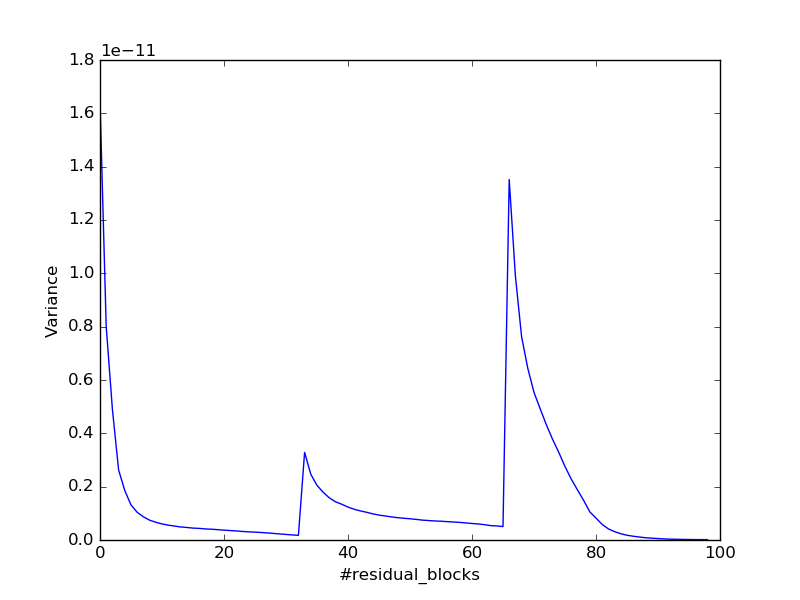}
\caption{The mean of the gradient variance as a function of the residual
block position at Epoch 1 (left), Epoch 25000 (middle) and Epoch 50000
(right) in case of resnet-99.}\label{fig:plot_164}
\end{figure*}

We observed similar results in case of Resnet-99, where the number of residual networks in each scale is 33. The results are shown in Fig(\ref{fig:plot_164}).

\section{Batch normalization vs residual branch in resnet}\label{sec:batchvsres}
We analysed the importance of batch normalization in resnet in the previous section. However, there is one question that needs to be solved, the relative importance of batch normalization w.r.t. the residual branches. We compared two variations of residual network with the original model. The models were trained on CIFAR-10. The models compared were 
\begin{itemize}
\item \textbf{Model-1:} Residual network with BN and residual branches
\item \textbf{Model-2:} Residual network with BN only (residual branches have been removed)
\item \textbf{Model-3:} Residual network with residual branches only (BN layers have been removed)
\end{itemize}

\begin{figure}[!ht]
\centering
\includegraphics[width=0.5\textwidth]{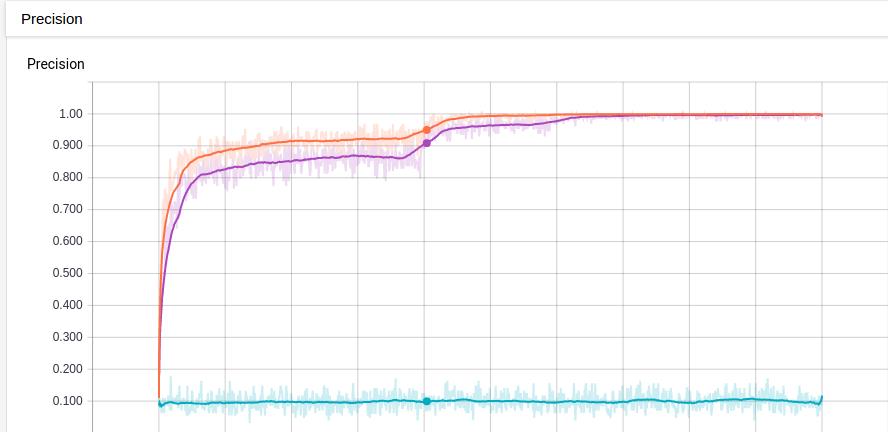}
\caption{Comparison of training accuracy for Model-1(red), Model-2(violet) and
Model-3 (blue).}\label{fig:train_Accuracy}
\end{figure}

All the models had 15 residual blocks, 5 in each scale. The parameters of each model were initialized similarly and were trained for same number of epochs. The weights were initialized with xavier initialization and the biases were initialized to 0.
First, we compare the training accuracy among the three models in  Fig.  \ref{fig:train_Accuracy}, where the horizontal axis shows the epoch number. We see that Model-1 reaches higher accuracy faster than the other two models. However, Model-2 isn't far behind. But Model-3, which has BN removed, doesn't learn anything. Next, we compare their test set accuracy in Table
\ref{table:2}. We see that Model-1 has the best performance while Model-2 isn't far behind.  

%%%%%%%%%%%%%%%%%%%%%%%%%%%%%%%%%%%%%  
\begin{table}[!ht]
\begin{center}
 \begin{tabular}{||c c||} \hline
 Model & Final accuracy\\ [0.5ex]  \hline\hline
 Model-1 & 92.5\% \\ \hline
 Model-2 & 90.6\% \\ \hline
 Model-3  & 9.09\% \\ [1ex] 
 \hline
\end{tabular}
\caption{Comparison of test accuracy of three Resnet models.}\label{table:2}
\end{center}
\end{table}
%%%%%%%%%%%%%%%%%%%%%%%%%%%%%%%%%%%%%  

\begin{figure}[!htb]
\centering
\includegraphics[width=0.33\linewidth]{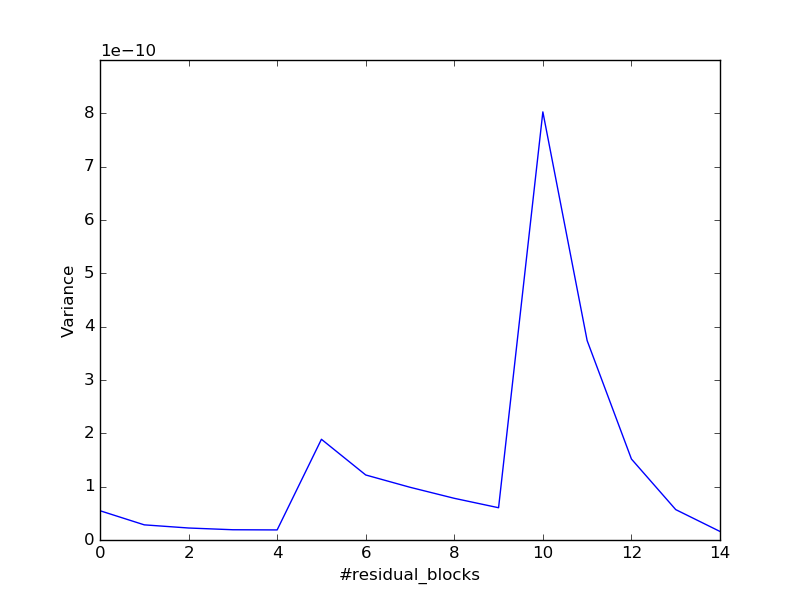}
\includegraphics[width=0.33\linewidth]{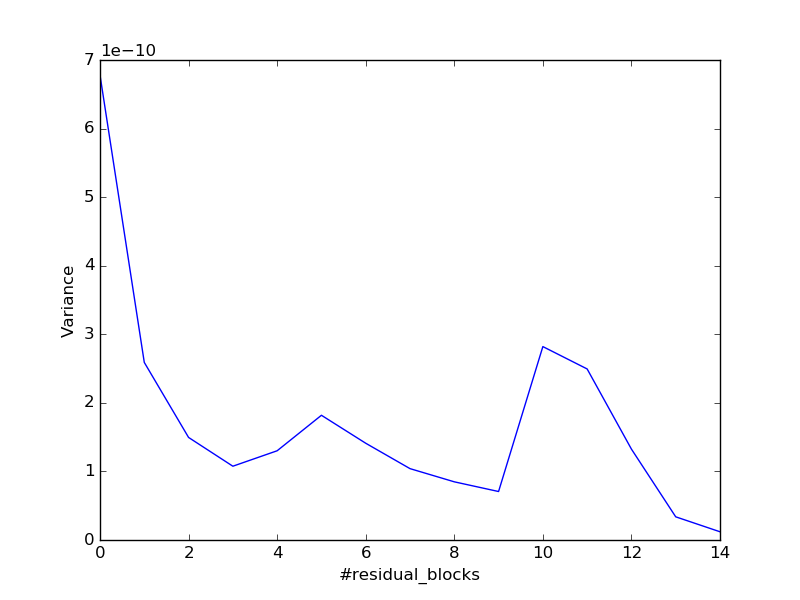}
\caption{The gradient variance as a function of the residual block index
during backpropagation in Model-1 (left), and
Model-2 (right) at Epoch 25000.}\label{fig:epoch_append-25000}
\end{figure}

\begin{figure*}[!htb]
\centering
\includegraphics[width=0.33\linewidth]{var_50000.png}
\includegraphics[width=0.33\linewidth]{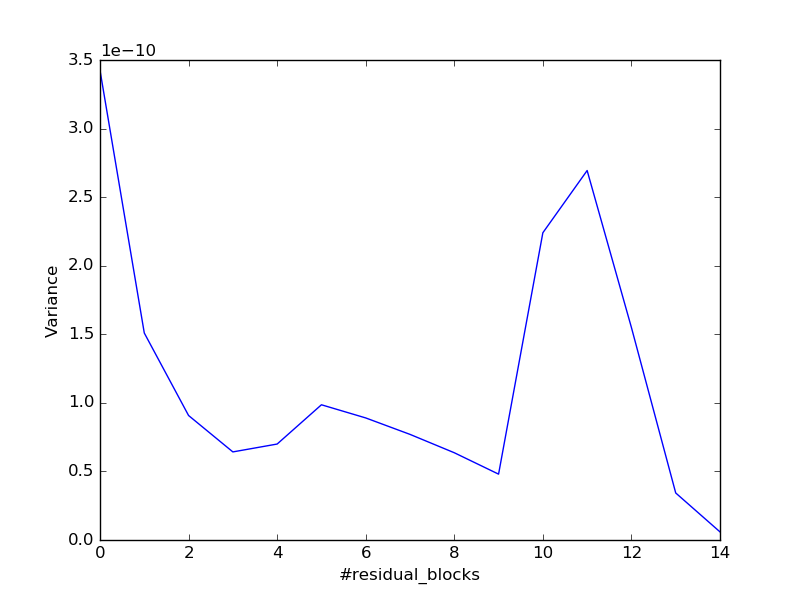}
\caption{The gradient variance as a function of the residual block index
during backpropagation in Model-1 (left), and
Model-2 (right) at Epoch 50000.}\label{fig:epoch_append-50000}
\end{figure*}

\begin{figure*}[!htb]
\centering
\includegraphics[width=0.33\linewidth]{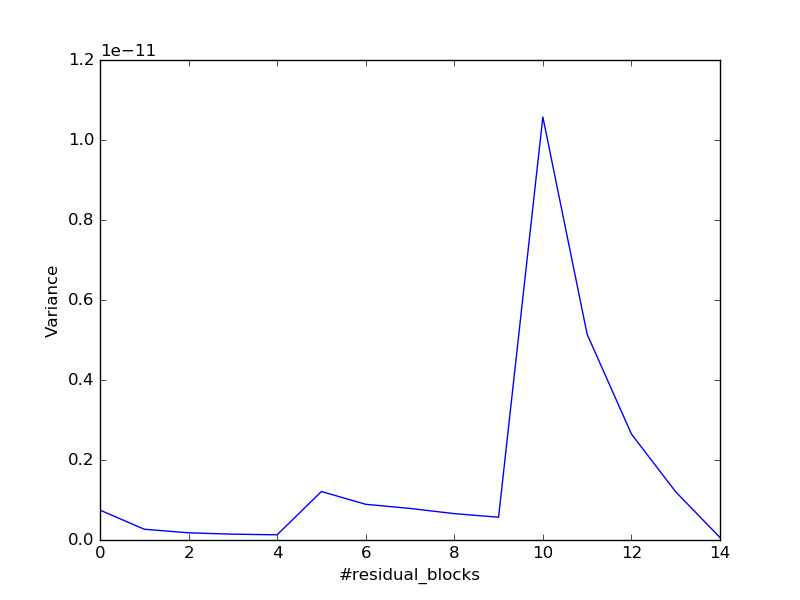}
\includegraphics[width=0.33\linewidth]{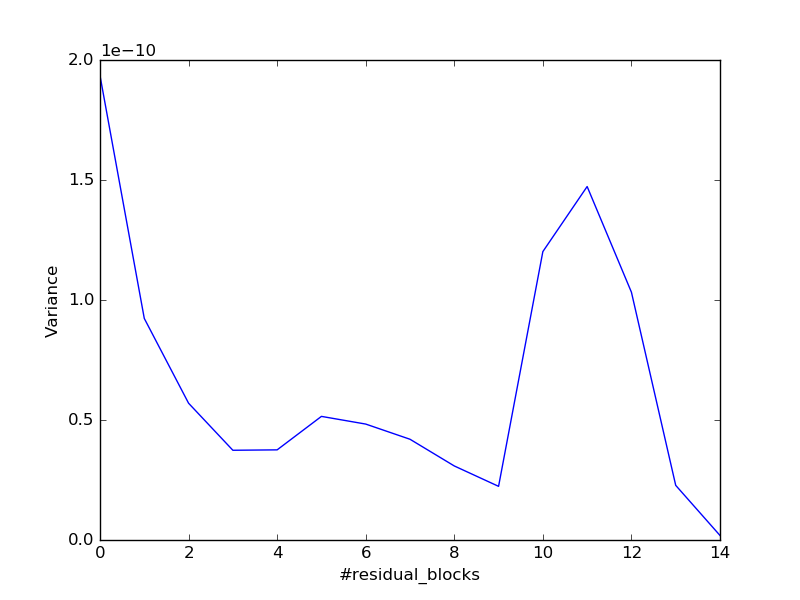}
\caption{The gradient variance as a function of the residual block index
during backpropagation in Model-1 (left), and
Model-2 (right) at Epoch 75000.}\label{fig:epoch_append-75000}
\end{figure*}

\begin{figure*}[!htb]
\centering
\includegraphics[width=0.5\linewidth]{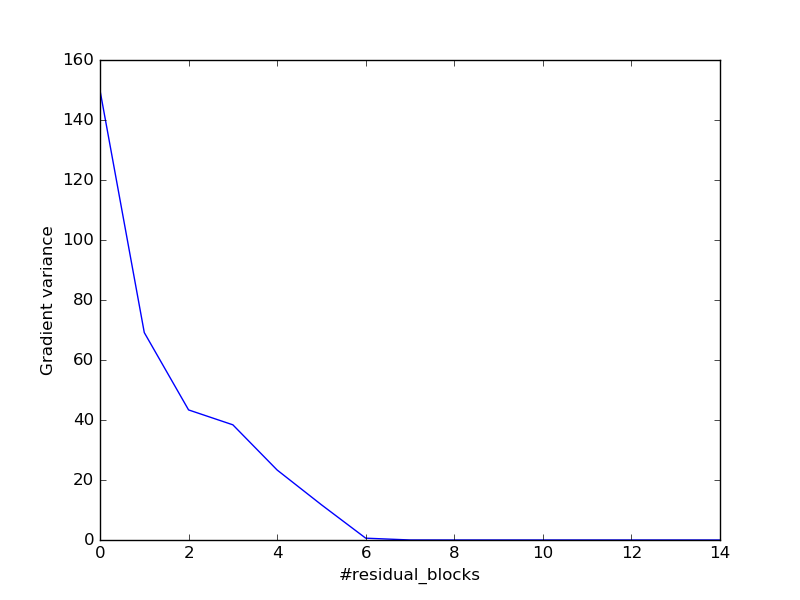}
\caption{Gradient explosion observed during back propagation in Model-3 at epoch-1}\label{fig:Model3}
\end{figure*}

Furthermore, we plot the mean of the gradient variance, calculated for each feature across one batch, as a function of the residual block index at epochs 25,000, 50,000 and 75,000 in Figs.
\ref{fig:epoch_append-25000}, \ref{fig:epoch_append-50000} and \ref{fig:epoch_append-75000},
respectively, where the performance of Model-1 and Model-2 is compared. We observe that 
the gradient variance also stays within a certain range, without exploding or vanishing, in 
case of Model-2. However, the change in gradient variance across a scale doesn't follow a fixed pattern compared to Model-1.
We also plot a similar kind of plot for Model-3 at epoch-1 in Fig \ref{fig:Model3}. We observed gradient explosion, right from the start, in case of Model-3 and the loss function had quickly become undefined. This was the reason, why Model-3 didn't learn much during the course of training. %Thus, we can see that batch normalization helps to stop gradient vanishing and explosion throughout training, thus stabilizing optimization.

This experiment shows that BN plays a major role in stabilizing training of residual networks. Even though we remove the residual branches, the network still tries to learn from the training set, with its gradient fixed in a range across layers. However, removing BN hampers the training process right from the start. Thus, we can see that batch normalization helps to stop gradient vanishing and explosion throughout training, thus stabilizing optimization.  Thus, the success of residual networks can't be attributed only to residual branches. Both BN and residual branches are responsible for the success of residual networks.

\color{black}

%%%%%%%%%%%%%%%%%%%%%%%%%%%%%%%%%%%%%%%%
%%%%%%%%%%%%%%%%%%%%%%%%%%%%%%%%%%%%%%%%

%%%%%%%%%%%%%%%%%%%%%%%%%%%%%%%%%%%%%%%%
%%%%%%%%%%%%%%%%%%%%%%%%%%%%%%%%%%%%%%%%

%%%%%%%%%%%%%%%%%%%%%%%%%%%%%%%%%%%%%%%%
%%%%%%%%%%%%%%%%%%%%%%%%%%%%%%%%%%%%%%%%

\section{Conclusion and Future Work}\label{sec:conclusion}

Batch normalization (BN) is critical to the training of deep residual
networks.  Mathematical analysis was conducted to analyze the BN effect
on gradient propagation in residual network training in this work.  We
explained how BN and residual branches work together to maintain
gradient stability across residual blocks in back propagation.  As a
result, the gradient does not explode or vanish in backpropagation
throughout the whole training process. We also showed experimentally the relative importance of batch normalization w.r.t the residual branches and showed that BN is important for stopping gradient explosion during training.

The Saak transform has been recently introduced by \citet{kuo2017saak},
which provides a brand new angle to examine deep learning.  The most
unique characteristics of the Saak transform approach is that neither
data labels nor backpropagation is needed in training the filter
weights. It is interesting to study the relationship between multi-stage
Saak transforms and residual networks and compare their performance in the near future.

%\newpage
\section{References}
\bibliographystyle{elsarticle-num-names.bst}
\bibliography{elsarticle-template-harv}

\newpage
\section*{Appendix A}

We apply the ReLU to the output of a BN layer, and show the step-by-step
procedure in calculating the variance and the mean of the output of the
ReLU operation. In the following derivation, we drop the layer and the
element subscripts (i.e., $L$ and $i$) since there is no confusion. Let $a=\beta/\gamma$.
\iffalse 
%%%%%%%%%%%%%%%%%%%%%%%%%%%%%%%%%
It
is assumed that scaling factors, $\beta$ and $\gamma$, in the BN are
related such that $a=\beta/\gamma$ is a small number and the
standard normal variable $z$ has a nearly uniform distribution in ($-a$,0). 
\fi
%%%%%%%%%%%%%%%%%%%%%%%%%%%%%%%%%%
Then,
we can write the shifted Gaussian variate due to the BN operation as
\begin{equation}
\gamma z + \beta = \gamma (z + a).
\end{equation}
Let $y = ReLU(z + a)$. Let a $>$ 0. We can write
\begin{equation}\label{eqn:Ey}
\begin{aligned}
E(y) =& P(z<-a) E(y|z<-a) + P(-a<z<0) E(y|-a<z<0)\\& + P(z>0) E(y|z > 0).
\end{aligned}
\end{equation}
The first right-hand-side (RHS) term of Eq. (\ref{eqn:Ey}) is zero since
$y=0$ if $z<-a$ due to the ReLU operation. Thus, $E(y|z<-a) = 0$. For the
second RHS term, we have%$z$ is uniformly distributed with probability density function equal to ${a}^{-1}$ in range (-a, 0) if $0< a << 1$. Then, we have
\iffalse
%%%%%%%%%%%%%%%%%%%%%%%%%%%%%%%%%%
\begin{eqnarray}
P(-a<z<0) = \frac{a}{\sqrt{2\pi}}, \quad \mbox{and} \quad E(y|-a<z<0)  = \frac{a}{2}.
\end{eqnarray}
%%%%%%%%%%%%%%%%%%%%%%%%%%%%%%%%%%
\fi

$$
P(0<z<a) = \int_{0}^{a} \frac{1}{\sqrt{2 \pi}} \exp(\frac{-z^2}{2})dz
$$

\begin{eqnarray}
E(y; -a<z<0) = \int_{0}^{a} z \frac{1}{\sqrt{2 \pi}} \exp(\frac{-z^2}{2})dz =  \frac{1}{\sqrt{2 \pi}} (1 - \exp(\frac{-a^2}{2}))
\end{eqnarray}

For the third RHS term, $P(z>0) = 0.5$. Besides, $z>0$ is
half-normal distributed.  Thus, we have
\begin{equation}
E(y| z > 0) = E(|z|) + a = \sqrt{\frac{2}{\pi}} +  a
\end{equation} 
Based on the above results, we get
\begin{equation}\label{eq:EY1}
E(y) = \frac{1}{\sqrt{2\pi}} + \frac{a}{2} + \frac{1}{\sqrt{2 \pi}} (1 - \exp(\frac{-a^2}{2}))
\end{equation}
Similarly, we can derive a formula for $E(y^2)$ as
\begin{eqnarray}
E(y^2) & = & P(z<-a)E(y^2|z<-a) + P(-a<z<0)E(y^2|-a<z<0) \nonumber \\
       & & + P(z>0)E(y^2| 0<z<a). \label{eq:Ey2}
\end{eqnarray}
For the first RHS term of Eq. (\ref{eq:Ey2}), we have
$E(y^2|z<-a)=0$ due to the ReLU operation.  For the second
RHS term of Eq. (\ref{eq:Ey2}), 
\iffalse
%%%%%%%%%%%%%%%%%%%%%%%%%%%%%%%%%%
$z$ is uniformly distributed
with probability density function ${a}^{-1}$ for -a$<$z$<$0 so that $P(-a<z<0) = \frac{a}{\sqrt{2\pi}}$ and
%%%%%%%%%%%%%%%%%%%%%%%%%%%%%%%%%%
\fi
$$
\label{true_var}
E(y^2; -a<z<0) = \int_{0}^{a} z^2 \frac{1}{\sqrt{2 \pi}} \exp(\frac{-z^2}{2})dz = \frac{-a}{\sqrt{2 \pi}} \exp(\frac{-a^2}{2}) -  \int_{0}^{a} \frac{1}{\sqrt{2 \pi}} \exp(\frac{-z^2}{2})dz
$$

For the third RHS term $P(z>0) = 0.5$ for 
$z>0$. The random variable $z>0$ is half normal distributed so that
\begin{eqnarray}
\begin{aligned}
E(y^2|z>0) =& E((|z|+a)^2) = E(|z|^2) + a^2 + 2aE(|z|) \\&= a^2 + 2\sqrt{\frac{2}{\pi}}a + 1.
\end{aligned}
\end{eqnarray}

Then, we obtain
%%%%%%%%%%%%%%%%%%%%%%%%%%%%%%%%%%
\iffalse
\begin{eqnarray}\label{eq:EY2}
E(y^2) = 0.5 + \sqrt{\frac{2}{\pi}}a + 0.5 a^2 + \frac{a^3}{3\sqrt{2\pi}}
\end{eqnarray}
\fi
%%%%%%%%%%%%%%%%%%%%%%%%%%%%%%%%%%
\begin{eqnarray}\label{eq:EY2}
E(y^2) = 0.5 + \sqrt{\frac{2}{\pi}} a + 0.5 a^2+\exp(\frac{-a^2}{2}) + p(a)
\end{eqnarray}

where p(a) =  $\int_{0}^{a} \frac{1}{\sqrt{2 \pi}} \exp(\frac{-z^2}{2})dz$.

\color{black}
We can follow the same procedure for $a<0$. The final results are summarized below.
\begin{equation}
E(ReLU (\gamma z + \beta) ) = \gamma E(y) \quad \mbox{and} \quad
E((ReLU(\gamma z + \beta ))^{2}) = \gamma^{2} E(y^2),
\end{equation}
where $E(y)$ and $E(y^2)$ are given in Eqs. (\ref{eq:EY1}) and (\ref{eq:EY2}), 
respectively.

\section*{Appendix B}

\begin{itemize}
\item We assumed that the function(F) applied by ReLU to the gradient vector and the gradient elements are independent of each other. Function F is defined as
$$
F(\Delta{y}) = \Delta{y} I_{y>0}
$$
where $\Delta{y}$ denotes input gradient in gradient backpropagation and y denotes the input activation during forward pass to the ReLU layer. $I_{y>0}$ = 1 when y$>$0 and it is 0 otherwise. Coming back to our analysis, since $\tilde{y}_{L-1, i}$ is a normal variate shifted by $a$, the probability that the input in forward pass to the ReLU layer, i.e. $\tilde{y}_{L-1, i}$ is greater than 0 is 
$$
P(\tilde{y}_{L-1, i} > 0) = 0.5 + p(a).
$$

where p(a) = $\int_{0}^{a} \frac{1}{\sqrt{2 \pi}} \exp(\frac{-z^2}{2})dz$.

\color{black}
Thus, E(F($\Delta{\hat{y}}_{L-1,i}$)) = E($\Delta{\hat{y}}_{L-1,i}$) P($\tilde{y}_{L-1, i} > 0$), and so
$$
E(\Delta{\tilde{y}}_{L-1,i}) = (0.5 + p(a)) \mbox{ } E(\Delta{\hat{y}}_{L-1,i})
$$
Similarly, we can solve for Var($\Delta{\tilde{y}}_{L-1,i}$) and thus, get Eq. (\ref{eq:Relu_grad}).

\item First, using eq \ref{eq:BN_only} and the assumption that the input standard normal variate in forward pass and the input gradient in gradient pass are independent, we have
\begin{eqnarray}\label{eq:30delta}
Var(\Delta y_{L-1,i}) & = & \frac{\gamma_i^2}{Var(y_{L-1,i})} Var(\Delta{\tilde{y}_{L-1,i}}) \\
& = &  \frac{\gamma_i^2}{Var(y_{L-1,i})} (0.5 + p(a)) \sum_{j=1}^{n_{L}^{\prime}} W_{L, ji}^{2} Var(\Delta y_{L, j}).
\end{eqnarray}

where p(a) = $\int_{0}^{a} \frac{1}{\sqrt{2 \pi}} \exp(\frac{-z^2}{2})dz$.

\color{black}
Then, using Eq. (\ref{eq:variance}) for $Y_{L-1}$ (yet with $L$ replaced with $L-1$), we 
can get Eq. (\ref{eq:gradient}).
\end{itemize}

\section*{Appendix C}
For $L = 1$, $\overline{y}_{1}= F (y_{0})$. Let $I_x$ denote a one vector of dimension x. Since the receptive field for the last scale is $k$ times smaller, we get the
following from Eq. (\ref{eq:variance}),
\begin{equation}
Var(\overline{y}_{1}) = c_2 \overline{W}_{1}^{2} I_{n_{1}}%\frac{n_{s}}{k} Var(\overline{W}_{1, i}^{f}).
\end{equation}
Also, since $\hat{y}_{1} = F(F(y_{0}))$, we have
$$
Var(\hat{y}_{1}) = c_2 \hat{W}_{1}^{2} I_{n_{1}}%\mbox{ } \frac{n_{s}}{k} \mbox{ } Var(\hat{W}_{1,i}^{f})
$$
based on Eq. (\ref{eq:variance}). Therefore, we get
\begin{eqnarray}
Var(y_{1}) & = & c_2 (\overline{W}_{1}^{2} I_{n_{1}} + \hat{W}_{1}^{2} I_{n_{1}})%\frac{n_{s}}{k} Var(\overline{W}_{1,i}^{f}) + 
%c_2 \frac{n_{s}}{k} Var(\hat{W}_{1,i}^{f}) \\
%& = & c_2  \frac{n_{s}}{k} ( Var(\overline{W}_{1,i}^{f}) + Var(\hat{W}_{1,i}^{f}) ).
\end{eqnarray}

For $L=N>1$, the input just passes through the shortcut branch. Then,
$$
Var(\overline{y}_{N,i}) = Var(y_{N-1, i}) 
$$
Also, since $\hat{y}_{N}= F( F( y_{N-1} ))$, we have 
$$
Var(\hat{y}_{N}) = c_2 \hat{W}_{N}^{2} I_{n_{N-1}}%\mbox{ } n_{s} \mbox{ } Var(\hat{W}_{N, i}^{f}).
$$
due to using Eq. (\ref{eq:variance}). Thus,
\begin{equation}
Var(y_{N}) = Var(y_{N-1}) + c_2 \hat{W}_{N}^{2} I_{n_{N-1}}%c_2 \mbox{ } n_{s} \mbox{ } Var(\hat{W}_{N, i}^{f}).
\end{equation}

Doing this recursively from $L=1$ to $N$, we get
\begin{equation}
Var(y_{N}) = c_2 (\sum_{J=2}^{N} \hat{W}_{J}^{2} I_{n_{J}} + \overline{W}_{1}^{2} I_{n_{1}} + \hat{W}_{1}^{2} I_{n_{1}})%n_{s} (\sum_{J=2}^{N} 
%Var(\hat{W}_{J,i}^{f}) + \frac{1}{k} ( Var(\overline{W}_{1, i}^{f}) 
%+ Var(\hat{W}_{1, i}^{f}) )).
\end{equation}

\section*{Appendix D}
Let $\tilde{K}_{L}$ and $\hat{K}_{L}$ denote the necessary upperbound for the convolution layer weights in the convolution branch, as needed by eq(\ref{eq:upperbound}).

For block $L = N>1$, the gradient has to pass through two
BN-ReLU-Conv Layers in convolution branch. Since, the receptive field doesn't change
in between the two BN-ReLU-Conv Layers in the convolution branch of the block, we use
Eq. (\ref{eq:gradient}) and find that for same receptive field between
the two layers i.e. $n_{L}$ $=$ $n_{L-1}^{\prime}$ ,
\begin{equation}
E(Var(\Delta{\tilde{y}_{L, i}})) \leq \hat{K}_{L} \frac{c_1}{c_2}
\frac{Var(\hat{W}_{L, .})}{Var(\tilde{W}_{L, .})}
E(Var(\Delta{y}_{L ,i})).
\end{equation}

When gradient passes through the first BN-ReLU-Conv Layer, the variance of the
forward activation that BN component sees is actually the variance of
the output of previous block. Hence, using Var($y_{L-1,i}$), which is the output of previous residual block, in place of the denominator in Eq. (\ref{eq:30delta}), we get
%$$
%Var(\Delta y_{L ,i}) = \frac{\gamma^2}{Var(y_{L-1},i)} Var(\tilde{y}_{L,i})
%$$
\begin{equation}
\begin{aligned}
E(Var(\Delta_{L ,i}))  \leq& \tilde{K}_{L} \frac{c_1}{c_2} \frac{Var(\tilde{W}_{L, .})}{\sum_{J=2}^{L-1} 
Var(\hat{W}_{J, .}) +  \frac{1}{k} ( Var(\overline{W}_{1, .}) + Var( \hat{W}_{1, .}) )} \\&
E(Var(\Delta{\tilde{y}_{L,i}})) 
\end{aligned}
\end{equation}

We assume that $\hat{\Delta}_{L}$ and $\Delta_{L}$ are independent of
each other. Since we are calculating for Block L$>$1 where there is no
BN-ReLU-Conv Layer in shortcut branch, we have $Var(\hat{\Delta}_{L,i}) =
Var(\Delta{y_{L, i}})$. As, 
$$
E(Var(\Delta{y_{L-1, i}})) = E(Var(\Delta_{L ,i})) + E(Var(\hat{\Delta}_{L,i})).
$$
Finally, we obtain
{\small
\begin{equation}
\begin{aligned}
E(Var(\Delta{y_{L-1,i}})) \leq& K_{L} \left( 1 + (\frac{c_1}{c_2})^2  
\frac{Var(\hat{W}_{L, .})}{\sum_{J=2}^{L-1} 
Var(\hat{W}_{J, .}) +  \frac{1}{k} ( Var(\overline{W}_{1, .}) 
+ Var(\hat{W}_{1, .}) )} \right)\\& E(Var(\Delta{y_{L, i}})).
\end{aligned}
\end{equation}
}%
where $K_{L} = \hat{K}_{L} \tilde{K}_{L}$.

\section*{Appendix E}
Here, we show the proof of
$$
E(\frac{1}{X}) E(X) \leq \frac{(c+d)^2}{4cd}
$$
where X is a variable and lies in the range (c,d), $0 < c < d$. 

The line $\frac{-1}{cd}X + \frac{c+d}{cd}$ intersects the curve $\frac{1}{X}$ at c and d. Hence, 
$$
E(\frac{1}{X}) \leq \frac{-1}{cd}E(X) + \frac{c+d}{cd} 
$$
$$ (E(\frac{1}{X}) + \frac{1}{cd}E(X))^2 \leq (\frac{c+d}{cd})^2
$$
But $(E(\frac{1}{X}) + \frac{1}{cd}E(X))^2 \geq 4 (E(\frac{1}{X})\frac{1}{cd}E(X))^2$ 
Finally, we get 
$$
4 (E(\frac{1}{X})\frac{1}{cd}E(X))^2 \leq (\frac{c+d}{cd})^2
$$
Since, d$>$c$>$0, we get
$$
E(\frac{1}{X}) E(X) \leq \frac{(c+d)^2}{4cd}
$$

\section*{Appendix F}

In this section, we show the experimental value of a = $\frac{\beta}{\gamma}$ during the training of residual networks containing 15 residual units. The mean of a (absolute value) at each layer stays reasonably small, atmost 1.0 in our case. This supports our theory that the constant term present in the upper and lower bounds of equation \ref{eq:upperbound} are reasonably small constants so that our theory holds true.

\begin{figure*}[!htb]
\includegraphics[width=0.32\linewidth]{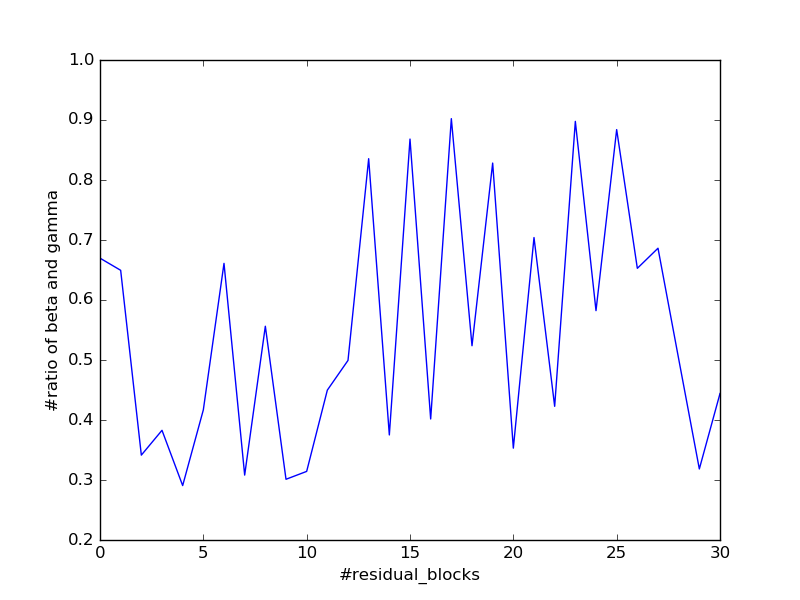}
\includegraphics[width=0.32\linewidth]{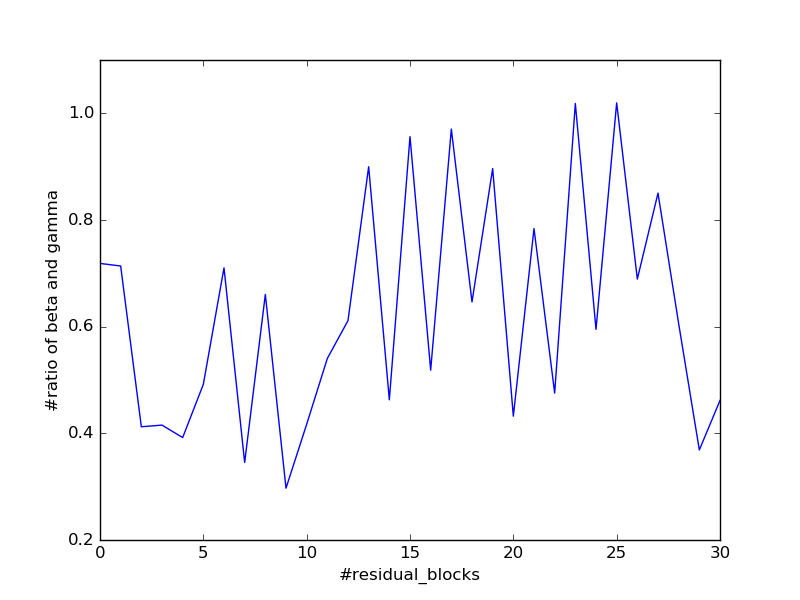}
\includegraphics[width=0.32\linewidth]{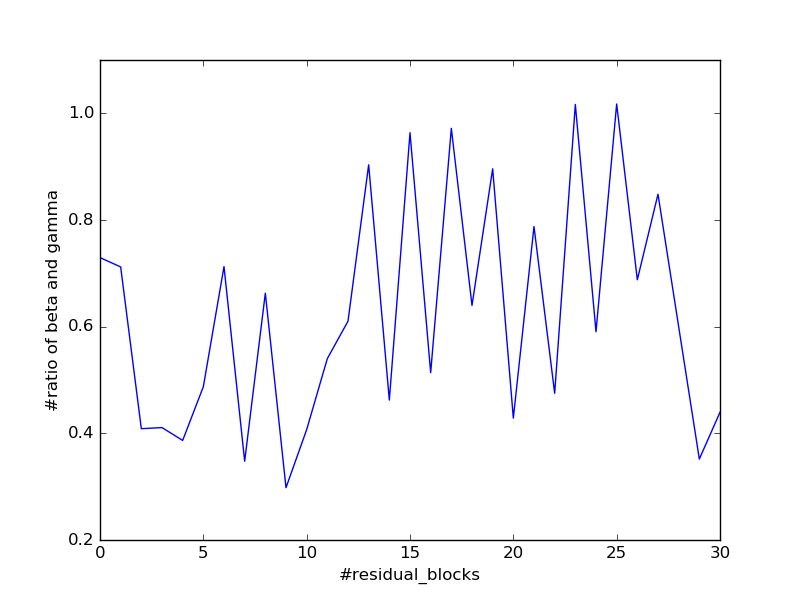}
\caption{The average  of absolute value of $\frac{\beta}{\gamma}$, calculated for each layer, as a function of the residual
block position at Epoch 25000 (left), Epoch 50000 (middle) and Epoch 75000
(right) in case of resnet-15.}\label{fig:beta_164}
\end{figure*}
\color{black}

\section*{Appendix G}

In this section, we empirically show that our theory stays independent of the batch size used in stochastic gradient descent and also the initialization of weights in the network. Equation \ref{eq:upperbound} is independent of the batch size used. Also, the initial variance of weights (within reasonable limits) doesn't affect the behavior of gradient variance during training.

\begin{figure*}
\includegraphics[width=0.32\linewidth]{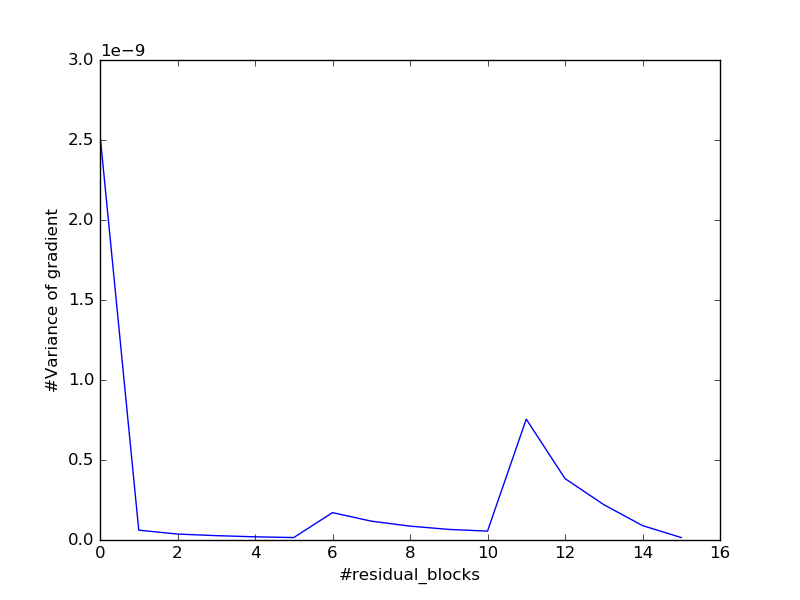}\hfill
\includegraphics[width=0.32\linewidth]{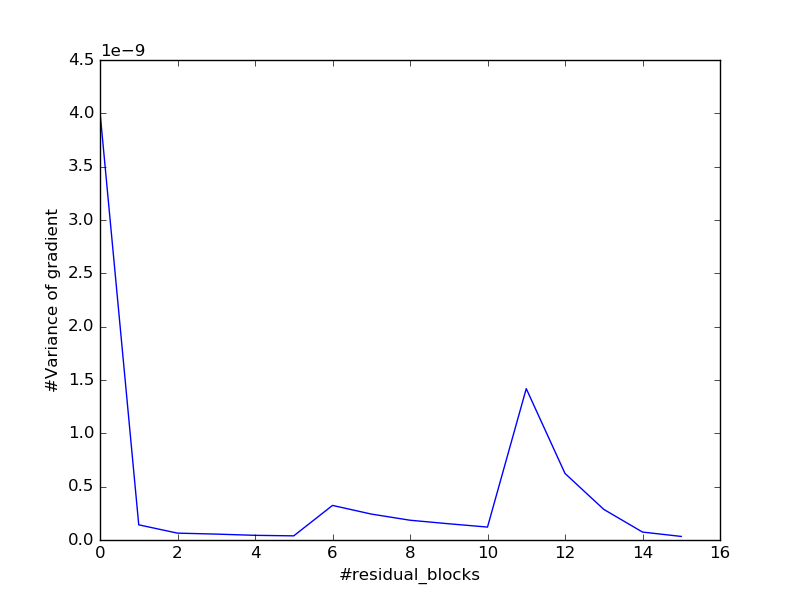}
\caption{The variance of gradient as a function of residual block measured at step 25000, when the weights are initialized by 0.01(left) and 0.1(right) respectively.  }\label{fig:plot_166}
\end{figure*}

\end{document}